\documentclass[11pt,letterpaper]{article}
\usepackage[margin=1in]{geometry}
\usepackage{amsmath,booktabs,graphicx}
\usepackage[hidelinks]{hyperref}
\newcommand{\benchmarkfile}[2]{\href{https://github.com/yuanrui3/Football-chess/blob/8e87600e949d23675a293414391c54e63cbc64aa/benchmarks/paired-positions/anc/#1}{#2}}
\hypersetup{pdftitle={When Should a Language Model Change Its Move? Paired Shot and Pass Decisions in Purdue Football Chess},pdfauthor={Yuanrui Chen}}
\title{When Should a Language Model Change Its Move?\\
Paired Shot and Pass Decisions in Purdue Football Chess}
\author{Yuanrui Chen\\
\small Elmore Family School of Electrical and Computer Engineering\\
\small Purdue University, West Lafayette, Indiana, USA}
\date{}
\begin{document}
\maketitle
\begin{abstract}
Which changes in a game state should change an agent's action? We use an exact engine to label both sides of 60 shot pairs and 40 pass pairs in Purdue Football Chess. Each pair differs by a one-square defensive edit. Half require switching actions. Across 1,080 shot decisions, Gemini 3.1 Flash-Lite solved 12 of 60 pairs from coordinates and rules, Gemma 4 solved three, and GPT-OSS-20B exhausted its generation budget on every prompt without reply odds. A simple shot-line rule solved 52 pairs. Supplying Gemini with conditional scoring odds raised its score to 57 pairs. Gemini changed shots on 33 pairs in either condition, but the number of required switches on which it changed rose from 16 of 30 to 30 of 30. On the original pass bank, Gemini and a tackle-range rule each solved 38 of 40 pairs. On eight fresh pairs, the author solved seven and Gemini solved two. Gemini chose the knight on 13 of 16 boards although it was optimal on four. In a further 48 text decisions, we supplied the possible tackle outcomes and varied the reliability of a scout report. Gemini made all 12 predicted reversals, six with the knight protected and six with the rook protected. With tackle outcomes supplied, Gemini switched receivers when the stated reliability crossed the protected pass's fixed 0.75 retention probability. How the model derives tackle exposure from a board remains open. We also calculate the minimum expected loss when an observation hides the edited piece, giving a reference for future tests under limited visibility.
\end{abstract}

\section{Introduction}
Suppose two lanes to goal are open. A defender takes one step. Has the best shot changed? We ask a model about each position independently, without showing its counterpart or identifying the edit. A model can get one position right by repeatedly favoring the same target. It can also choose different shots on the two positions and be wrong both times. A benchmark for decision-making needs to distinguish the \emph{frequency} with which a policy changes its action, the \emph{edits} on which it changes, and whether it selects the \emph{right action} on each side of the edit.

Purdue Football Chess makes these distinctions observable. In the shot task, the engine assigns an exact scoring probability to each target after every legal goalkeeper reply. The best shot can therefore be identified independently of human preferences or an estimate of who eventually wins a match. Both members of a pair offer the same three shots in the same order. A model receives only one board per request. The experimenter can compare its choices across the edit without telling it that anything changed.

Two information conditions led Gemini to change its shot on exactly 33 of 60 pairs, yet the two sets of changes had very different relationships to the true decision boundary. We compare its choices to a geometric policy and report three model families under controlled disclosure of goalkeeper moves and scoring odds. The original endpoint was the number of shot pairs solved as more information was supplied. The geometry policy and the subsequent analysis of action changes were exploratory. We then froze a second test on 40 pairs of passing positions. All three passes reach their receivers. Which one retains possession depends on whether a defender can tackle the recipient. This lets us test the same distinction using a different action, objective and spatial rule. A further eight pairs ask whether the model still finds a safe pass when the knight is usually a costly choice. We then vary the reliability of a report about an explicitly described tackle while swapping which piece has protection. On the three fully observed banks, the engine values also let us calculate how much an ideal observer would gain by resolving an otherwise hidden edit.

\paragraph{Related work.}
Contrast sets~\cite{gardner} evaluate decisions on nearby examples, and CheckList~\cite{ribeiro} distinguishes perturbations that should preserve outputs from those that should change them. GameBench~\cite{costarelli} assesses agents through competitive play across nine games. The MATE chess dataset~\cite{wang} presents candidate moves and studies the effect of explanations of strategy and tactics. LudoBench~\cite{jain} evaluates choices in stochastic board-game positions against a depth-limited expectiminimax agent and varies narrative context at fixed board states. Here the intervention changes one physical position in the game state. An exact one-decision objective labels both the optimal action and whether that edit crosses its decision boundary. Controlled access to the per-reply payoffs makes it possible to score the same decisions at distinct levels of information.

\section{Shot task and position pairs}
The pitch has 14 files and 17 ranks. Black attacks the goal below rank 1 and White attacks above rank 17. Appendix~\ref{app:rules} gives the rules relevant to both tasks, including the numerical shot model. A ball carrier near goal can declare either of two target files or disguise the shot, in which case the two targets are equally likely. A goalkeeper can hold or take one legal step, including diagonals. For a declaration the goalkeeper knows the target. For disguise it must choose one destination before learning the target. Let $q(r,t)$ be the engine probability of scoring at target $t$ after reply $r$, and let $R$ be the set of legal replies. The attacker maximizes over three action values:
\begin{equation}
 V(t_j)=\min_{r\in R} q(r,t_j),\quad j\in\{1,2\},\qquad
 V(D)=\min_{r\in R}\frac{q(r,t_1)+q(r,t_2)}{2}.
 \label{eq:value}
\end{equation}
Taking the mean within each reply matters. Averaging the two column minima would allow the keeper to react to a target that a disguise has yet to reveal.

\paragraph{Sampling and verification.}
Three fixed generation seeds (41711, 51329, 60509) each contributed ten pairs in which the best shot changes (\emph{switch}) and ten in which it stays optimal (\emph{stay}). One defending keeper or pawn moves one square horizontally or vertically. The attacking side, shooter, targets, other pieces and shot menu stay fixed. The generator required a gap of at least 1.5 percentage points between the two best actions on both boards. Every switch exchanged disguise and a declaration. Across the 120 boards each of the three menu positions was correct 40 times. We excluded repeated payoff tables within the bank and across four earlier project banks, accounting for target exchange, row order and duplicated equivalent rows. An independent implementation of the printed rules reproduced all 601 legal replies and 360 action values, with maximum absolute probability error below $5.1\times 10^{-16}$. Rounding the displayed reply probabilities to two decimal percentage points preserved every optimum.

\begin{figure}[htbp]
\centering
\includegraphics[width=.89\linewidth]{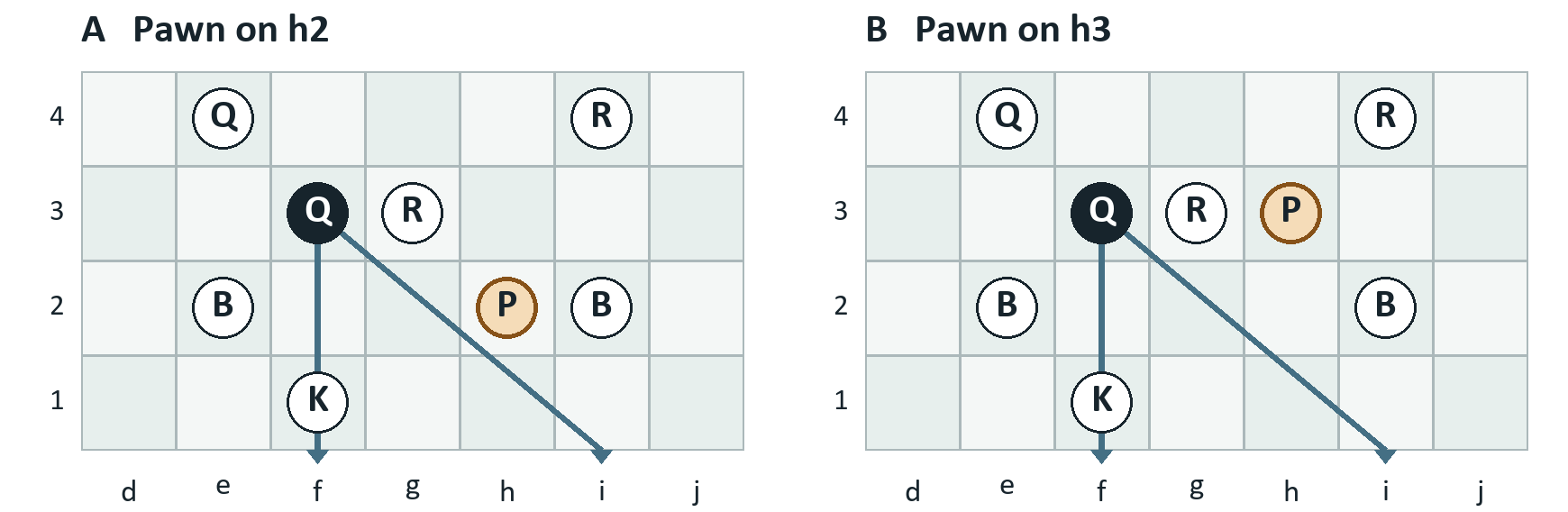}
\caption{Two goalmouths from pair \texttt{info-3-16}. Black's queen shoots from f3 toward f or i. Moving the highlighted white pawn from h2 to h3 clears the line to i. The keeper on f1 has the same five replies in both positions. The figure is for readers. Models received coordinates.}
\label{fig:pair}
\end{figure}

In Figure~\ref{fig:pair}, declaring f achieves 25.87\% against the best reply on both boards. Disguise rises from 24.04\% to 33.12\%. In the second position the keeper's best reply against disguise, f2, leaves goal probabilities of 25.87\% and 40.37\%, giving 33.12\% when averaged. The model must distinguish an increase in one target's value from a change in which action wins. In another pair (\texttt{info-1-04}), the same pawn step h2--h3 raises the value of disguise from 60.62\% to 70.13\%, but declaring i remains best at 98.00\% on both boards. These examples were selected after scoring to illustrate opposite outcomes of the same edit.

\section{Information conditions and policies}
Each model answered each shot position in three fresh contexts. \emph{Board} provided occupied squares, numerical rules and the menu. \emph{Replies} added all legal goalkeeper destinations. \emph{Odds} added both conditional scoring percentages to each destination. The odds condition still requires the minimax comparison in Equation~\ref{eq:value}. Each request used the same user text across models for a given board and condition. A shuffled schedule interleaved conditions and seeds. Models had no tools, feedback, counterpart, or pair identifier.

The hosted Gemini 3.1 Flash-Lite and Gemma 4 31B IT comparisons were frozen before collection in November 2026. After unscored availability failures for Gemma, a separate freeze specified a local GPT-OSS-20B comparison on the same inputs. These freezes were internal protocol snapshots recorded with timestamps and request and source hashes. Gemma subsequently completed its arm. Google developed and hosted both Gemini and Gemma. GPT-OSS ran in native MXFP4 with llama.cpp. Each model used its high-reasoning setting, temperature zero and a 4,096-token generation limit. The experiment collected one response per model, board and condition, or 1,080 primary decisions. Each completed invalid or truncated answer counted as incorrect. A transport failure was retried. An answer had to specify one legal menu option in a completed JSON object.

\paragraph{Geometry policy.}
The exploratory policy uses the public coordinates and enumerates goalkeeper holds and legal steps. For each reply and target it traces the bounded segment from the shooter to the target center, half a rank beyond the goal line. Let $m$ count opposing bodies whose square interiors intersect the segment. A knight ignores blockers. A keeper outside its box can block. For a keeper inside its box, let $d$ be its distance in squares to the closest point on the segment. The target score is
\[
 s(r,t)=0.55^m\,g(r,t),\qquad
 g(r,t)=\begin{cases}
1-\exp(-d^2),&\text{in-box keeper with }(K-S)\cdot(T-S)\geq 0,\\
 1,&\text{otherwise}.
 \end{cases}
\]
Here $K,S,T$ are the keeper, shooter and target centers. Substituting $s$ for $q$ in Equation~\ref{eq:value} gives three scores. Ties favor disguise. The blocker factor 0.55 comes from the printed rules. The approximation omits angle, range and keeper timing. No coefficients were fitted. The rule was specified after the primary freeze and before inspection of model scores.

\paragraph{Scoring.}
A pair is solved if both selected actions are optimal. Shot loss is the difference in scoring probability between the optimal and selected actions, averaged over valid decisions. Our exploratory action-change diagnostic compares the \emph{semantic} shots selected on the two boards. For each kind of pair we count changes among pairs with two valid choices, and fully solved pairs among all assigned pairs. A change in the wrong direction contributes to the first count only. Independent uniform choices from the three shots would change the selected action on 20 of 30 pairs in each group, in expectation.

\section{Shot results}
\begin{table}[htbp]
\centering
\caption{Action changes and correctness on the same 60 shot pairs. Each switch or stay column has denominator 30. All listed policies returned a valid choice on both boards of every pair.}
\label{tab:changes}
\begin{tabular}{lrrrr}
\toprule
&\multicolumn{2}{c}{Changed action}&\multicolumn{2}{c}{Both optimal}\\
\cmidrule(lr){2-3}\cmidrule(lr){4-5}
Policy and information & Switch & Stay & Switch & Stay\\
\midrule
Geometry, board &27&1&24&28\\
Gemini, board &16&17&3&9\\
Gemini, with replies &18&18&6&6\\
Gemini, with reply odds &30&3&30&27\\
\bottomrule
\end{tabular}
\end{table}

Gemini changed its shot on 33 pairs from coordinates and on 33 pairs with odds (Table~\ref{tab:changes}). On the board alone it changed on 16 of the 30 required switches and on 17 of the 30 stable pairs. With odds it changed on every required switch and on three stable pairs. This shift is invisible in the overall change rate. Even changing on a switch did not ensure a correct pair. Gemini solved only three of its 16 changed switch pairs from the board. Figure~\ref{fig:pair} illustrates one such failure. Gemini changed from declaring i to declaring f, whereas the optimal choices were declaring f and then disguise.

The geometry policy solved 52 pairs (16, 19 and 17 out of 20 across seeds). From the same board information, Gemini and geometry both solved 11 pairs, geometry alone solved 41, Gemini alone solved one and both missed seven. Geometry solved every keeper-move switch (21/21) and three of nine pawn-move switches. It chose disguise on both boards in Figure~\ref{fig:pair}. On the first board, the missed declaration cost 1.84 percentage points using the unrounded probabilities. The pair scored zero. Table~\ref{tab:policies} reports both the frequency and the consequence of errors.

\begin{table}[htbp]
\centering
\caption{Deterministic shot policies. Loss is the mean reduction in scoring probability in percentage points. The final row receives the displayed payoff table.}
\label{tab:policies}
\begin{tabular}{lrrrrr}
\toprule
Policy & Boards /120 & Pairs /60 & Switch /30 & Stay /30 & Loss\\
\midrule
Always disguise &38&4&0&4&14.86\\
Aim farther from keeper &71&27&4&23&7.84\\
Use reachable target files &73&29&6&23&7.74\\
Simplified shot geometry &110&52&24&28&0.68\\
Minimax on supplied odds &120&60&30&30&0.00\\
\bottomrule
\end{tabular}
\end{table}

The farther-file policy chooses the target with the larger horizontal distance to the keeper, disguising on a tie. The reachable-files policy picks the uncovered file when exactly one target file is reachable by an in-box keeper, disguises if both are reachable and otherwise uses the farther-file rule. Independent uniform guessing solves $60/9=6.67$ pairs in expectation.

\begin{table}[htbp]
\centering
\caption{Correct shot boards out of 120 and fully solved pairs out of 60. The first three rows used the primary protocol. The final row re-elicited all 360 local-model decisions in a separate follow-up.}
\label{tab:models}
\begin{tabular}{lrrr}
\toprule
 & Board & With replies & With reply odds\\
\midrule
Gemini 3.1 Flash-Lite &52 / 12&52 / 12&117 / 57\\
Gemma 4 31B IT &34 / 3&38 / 4&42 / 6\\
GPT-OSS-20B &0 / 0&0 / 0&83 / 30\\
GPT-OSS final-answer follow-up &37 / 6&38 / 4&90 / 35\\
\bottomrule
\end{tabular}
\end{table}

\paragraph{Prespecified comparisons.}
For each model, adjacent conditions were compared on discordant whole pairs by a two-sided exact binomial test. Each frozen two-model comparison adjusted four tests using Holm's procedure~\cite{holm}. Marginal 95\% percentile intervals used 20,000 bootstrap resamples of whole pairs within the three seed blocks. Adding legal replies to the board gained Gemini seven pairs and lost seven. Adding odds gained 46 and lost one, a 75.0 percentage-point increase (95\% interval 63.3 to 86.7, adjusted $p<0.001$). Mean shot loss fell from 14.22 to 0.59 points. For Gemma, odds gained two pairs and lost none (3.3 points, interval 0.0 to 8.3, adjusted $p=1$). Gemma returned 25 answers with a trailing Markdown fence. Removing that delimiter in a secondary check gave 4, 4 and 6 solved pairs.

\paragraph{Completion effects.}
GPT-OSS exhausted its generation budget without a choice on all 240 primary questions without odds and on 30 of 120 with odds. Adding odds gained 30 pairs and lost none (50.0 points, interval 38.3 to 63.3, adjusted $p<0.001$). To estimate its choices under a different elicitation procedure, a separately frozen post hoc follow-up repeated all 360 decisions. It allowed 2,048 tokens initially. For 304 responses that produced reasoning and an empty final answer, a continuation from that reasoning constrained the final answer to one of the three choices, within a combined ceiling of 4,096 tokens. All 360 choices were valid. Adding odds gained 32 pairs and lost one (51.7 points, interval 38.3 to 65.0). As Table~\ref{tab:models} shows, supplying odds still helped when every question received an answer. The two protocols also allocated reasoning tokens differently.

\section{A second decision: choosing a pass recipient}
The shot result raises a question beyond shooting. Does the same evaluation expose shortcuts when the agent must choose a teammate instead of a target file? We constructed a separate receiver-selection task after examining the shot study, and froze its boards, scoring protocol, and Gemini requests before the first pass response. The agent chooses among three legal, onside passes from a queen. Every offered pass reaches the receiver with probability one. The relevant geometry is at the destination, where a defender may win the ball on the next move. Appendix~\ref{app:passing} specifies the tackle rules.

\paragraph{Objective and boards.}
The attack holds its positioning move after the pass. The defending side may then hold or make one legal move. With fouls disabled, a legal tackle wins the ball with probability one against a non-knight and one-half against a knight. A player who cannot be tackled retains the ball. For receiver $j$ let $p_{\mathrm{win}}(d,j)$ be the probability that defensive move $d$ wins possession. The engine value of passing to $j$ is $1-\max_d p_{\mathrm{win}}(d,j)$. This objective fixes the horizon and the attack's subsequent policy. It makes no prediction about which pass will ultimately produce a goal.

Seed 20261127 produced 40 pairs, with ten in each combination of switch or stay and attacking color. Each position contains three candidates on distinct passing lines two to six squares from the carrier. One defender moves a single square, possibly diagonally, between the constructed alternatives. The three recipients and their menu order are fixed within each pair. Of 1,678 proposals the generator retained 40 pairs. A preliminary screen using tackle shapes, range and intervening pieces sought the required switch or stay, a unique optimum, and a change in at least one receiver's value. The native engine verified every retained value. This sampling favors positions that can be explained by a short geometric calculation. Across the 80 positions the three menu slots are optimal 27, 26 and 27 times. A second verification replayed the pass, the hold and all 1,640 legal tackle landings across the candidates, checking both outcomes where tackles were stochastic. The 80 Gemini prompts gave the complete rules and coordinates. Each was presented in its own context with the same hosted model settings used for the shot task.

\begin{figure}[!htb]
\centering
\includegraphics[width=.86\linewidth]{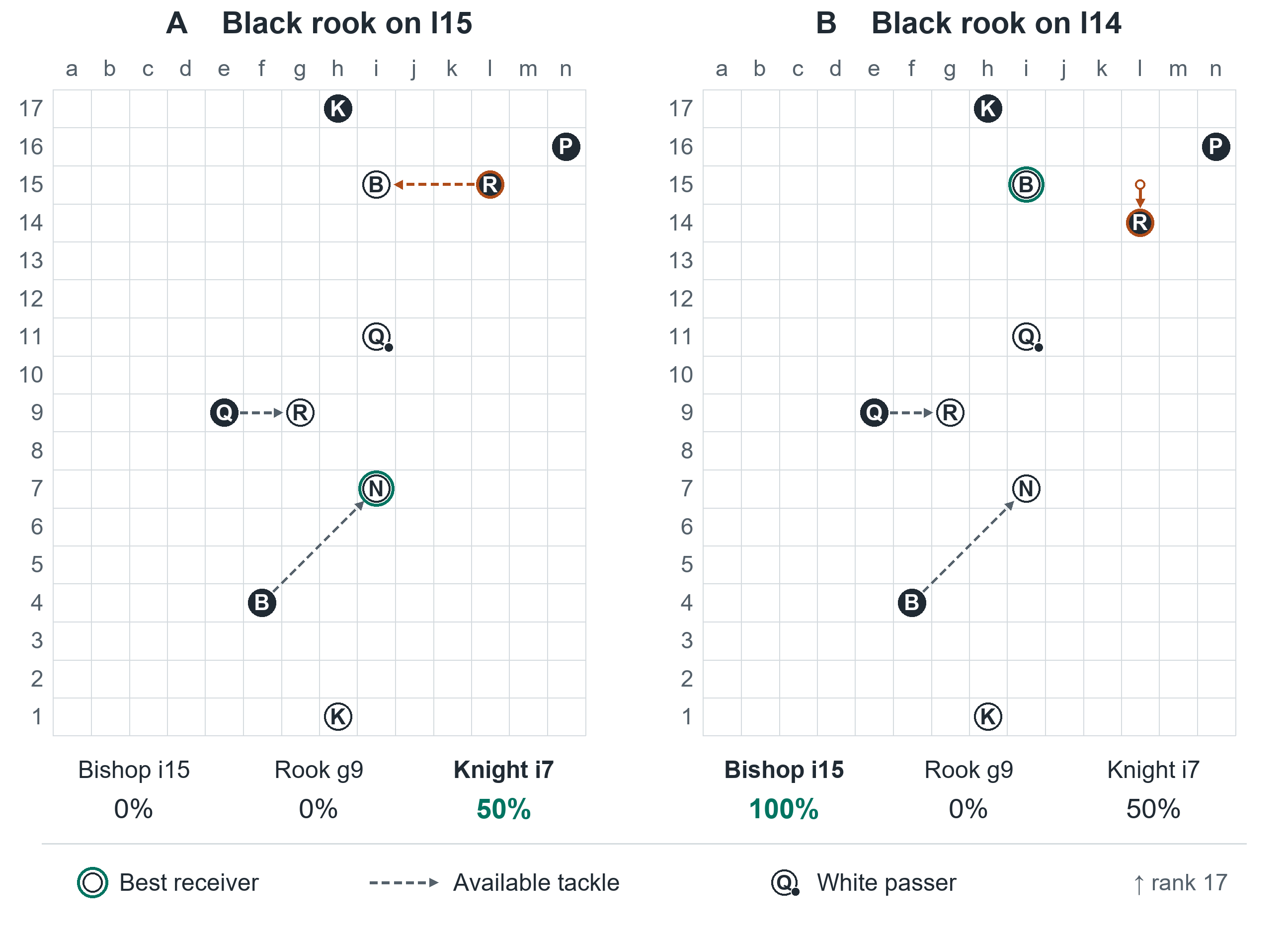}
\caption{Passing pair \texttt{p03}, chosen before inspecting model responses. White's queen on i11 can reach all three receivers. Dashed arrows mark tackles available after reception. Moving the highlighted black rook from l15 to l14 raises the white bishop's retention on i15 from zero to one. The white rook on g9 stays at zero and the knight on i7 at one-half. The double ring marks the best receiver. The two boards illustrate alternative arrangements; the edit need not be a legal move between successive turns.}
\label{fig:pass}
\end{figure}

\paragraph{What the pair score adds.}
The generator produced a tempting shortcut: on 78 of 80 boards a knight was among the candidates, and on 59 it was optimal. On 39 boards every receiver was exposed and only the knight could survive a tackle. All 20 stay pairs favor the knight twice. Nineteen switch pairs exchange a knight and a non-knight, and one exchanges a rook and a bishop. A post hoc policy that passes to the knight when offered and otherwise chooses the first receiver scores 60/80 boards. It solves all 20 stays and none of the 20 switches. The frozen geometry rule in Table~\ref{tab:pass} checks each defender's tackle shape and range, ignoring intervening pieces and the legality of a tackle landing beside the ball. It misses two boards. All deterministic policies break ties in menu order.

\begin{table}[htbp]
\centering
\caption{Passing decisions from coordinates. Each switch and stay group has 20 pairs. Loss is the mean decrease in possession probability in percentage points across valid choices. The knight rule was added after the bank was frozen.}
\label{tab:pass}
\begin{tabular}{lrrrrrr}
\toprule
& &\multicolumn{2}{c}{Changed}&\multicolumn{2}{c}{Both optimal}&\\
\cmidrule(lr){3-4}\cmidrule(lr){5-6}
Policy & Boards /80 & Switch & Stay & Switch & Stay & Loss\\
\midrule
First receiver &27&0&0&0&7&46.25\\
Farthest from a defender &37&4&1&3&9&37.50\\
Knight when offered &60&0&0&0&20&13.13\\
Tackle shapes and ranges &78&18&0&18&20&1.25\\
Gemini 3.1 Flash-Lite &78&19&0&18&20&1.88\\
\bottomrule
\end{tabular}
\end{table}

Gemini returned 80 valid choices. It changed receivers on 19 of 20 required switches and held its choice on every stay pair. It solved 38/40 pairs, the same total as the tackle-shape rule. They solved 36 pairs in common and two each that the other missed. Gemini's two errors cost an average of 1.875 percentage points across the 80 boards, compared with 1.25 for the geometry rule. Equal pair scores here describe two distinct sets of decisions.

Appendix~\ref{app:walkthrough} follows four of these pairs through a subsequent visual walkthrough by the author. On its two stay pairs, the best receiver remained the same even as its chance of retaining possession fell from one to one-half.

\subsection{A fresh test of receiver choice}
\label{sec:fresh}
The knight was optimal on 59 of the original 80 boards, including both members of every stay pair. We constructed eight fresh pairs to test receiver choice when an exposed knight usually had a safer alternative. Each menu contained a bishop, rook and knight. The knight retained possession with probability $1/2$ on all 16 boards. The rook and bishop were each uniquely optimal on six boards, the knight on four. A defending pawn moved one square within each pair. Four pairs placed a pawn between a sliding defender and a receiver on at least one board. On the other four, a rule using tackle shape and range alone recovered every receiver's value. Each group contained two switch and two stay pairs. The engine checked all 48 pass values across 328 legal tackle landings.

The bank, requests and primary comparison were frozen before responses were collected. The author and Gemini received the same pitch image, coordinate roster, rules and three options for each board. The quiz locked each confirmed choice, separated counterparts by at least five other boards and gave no feedback. The author had seen the earlier results and could remember earlier questions. Gemini received each board in a fresh context with the original temperature, reasoning setting and generation limit. It returned one option number. All 16 model choices were recorded before joint scoring.

\begin{table}[htbp]
\centering
\caption{Fresh passing decisions. Each of the four middle columns counts pairs with both receivers correct. Switch/stay and blocked-line/control are two partitions of the same eight pairs. Loss is mean possession probability forgone, in percentage points.}
\label{tab:fresh}
\small
\begin{tabular}{lrrrrrr}
\toprule
& Boards & Switch & Stay & Blocked-line & Control & Loss\\
& /16 & /4 & /4 & /4 & /4 & (pp)\\
\midrule
Author &15&4&3&3&4&3.125\\
Gemini 3.1 Flash-Lite &7&2&0&1&1&28.125\\
\bottomrule
\end{tabular}
\end{table}

On the prespecified comparison, the author solved seven of eight pairs and Gemini solved two (Table~\ref{tab:fresh}). Both solved two, the author alone solved five, and neither solved one. A post hoc count showed that Gemini chose the knight on 13/16 boards, including all nine errors. On each error it passed up a certain retention for a one-half chance. Its three other choices went to bishops. It never chose a rook. The author chose the knight on 5/16 boards. Gemini solved one blocked-line pair and one control pair, so its errors also occurred where ignoring intervening pieces preserved all three values. The comparison describes one participant familiar with the benchmark and one model run.

We also checked the subset of boards in each bank offering both an exposed knight and a uniquely safe ordinary receiver. Gemini chose the safe receiver on 18/19 such boards in the original bank and 3/12 in the follow-up. Every other choice on these boards went to the knight. The change in how often a knight was optimal cannot by itself account for this difference within a common decision type. This post hoc comparison changed three things together. The boards came from different generators, the follow-up added a pitch diagram, and the response format changed from a JSON object to one option number. Repeating the same boards while varying one of those inputs at a time would separate their effects.

\section{What is it worth to see the edit?}
\label{sec:information}
How much would an optimal decision-maker lose if the observation concealed which board was present? We calculate this quantity after collection by hiding the edited defender's square in both descriptions while retaining its identity and the rest of the board. This mask leaves every pair distinguishable from the other pairs in its bank. Consider a known pair of equally likely states $s_0,s_1$ with the same three actions. Write $V_i(a)$ for the engine value of action $a$ in state $s_i$, including the defender's best reply. Suppose the decision-maker knows both payoff vectors and initially gets the same observation in both states. The value of revealing the member before action is the classical value of perfect information~\cite{howard},
\begin{equation}
 G=\tfrac12\big[\max_a V_0(a)+\max_a V_1(a)\big]
     -\max_a\tfrac12\big[V_0(a)+V_1(a)\big].
 \label{eq:reveal}
\end{equation}
The first term allows a separate choice in each state. Without the reveal, a randomized choice averages the three common-action values and cannot exceed their maximum. This proves that $G$ is the smallest expected regret attainable from the indistinguishable observation. It vanishes exactly when the two states share an optimal action. A stay can therefore have zero reveal value even when an action's payoff changes substantially. Computing Equation~\ref{eq:reveal} across the original banks gives mean gains of 2.26 percentage points in scoring probability for shots and 13.125 points in possession for passes. Both banks contain equal numbers of switches and stays, yet the cost of missing the edit differs substantially.

In fresh pair \texttt{m02}, Black can pass to its rook on e11. A white pawn on g11 screens the rook from the white queen on i11. Moving the pawn to g10 opens the tackle and drops the rook's retention from 1 to 0. The knight on l10 stays at $1/2$ and the bishop at 0. With the pawn square hidden, either the rook or the knight averages $1/2$. Seeing the square lets the passer choose the rook when screened and the knight otherwise, averaging $3/4$. The author made this switch. Gemini chose the knight twice. All four fresh switch pairs have the same payoff pattern up to receiver identity, giving $G=1/4$ on each. The four stays give $G=0$. Across all eight equally weighted pairs the expected gain is $1/8$, or 12.5 percentage points of possession. The image and the coordinates supplied to both decision-makers in Section~\ref{sec:fresh} displayed the pawn.

\paragraph{A prediction for partial visibility.}
Consider any of these four switch pairs under the same prior. Suppose a signal $z$ has state-conditional distributions $P_0,P_1$. Let $s_1$ be the state in which the ordinary receiver is safe and $p(z)=\Pr(s_1\mid z)$. That receiver has expected retention $p(z)$, the exposed knight $1/2$, and the third receiver zero. The optimal policy chooses the ordinary receiver when $p(z)>1/2$, the knight when $p(z)<1/2$, and either on a tie. Among these two undominated actions, a wrong state classification costs exactly $1/2$ in either direction. For a discrete signal, the optimal classifier's error is $\tfrac12\sum_z\min\{P_0(z),P_1(z)\}$. Its expected loss in retention relative to perfect observation is therefore
\begin{equation}
 R^*=\tfrac14(1-\mathrm{TV}(P_0,P_1)),\qquad
 \mathrm{TV}(P_0,P_1)=\tfrac12\sum_z|P_0(z)-P_1(z)|.
 \label{eq:visibility}
\end{equation}
This gives a floor for a specified observation channel. For example, a binary cue reporting the correct pawn square with probability $0.8$ and the other square with probability $0.2$ in each state gives an optimal retention loss of 10 percentage points. On boards with the original one-half knight protection, a future test could vary cue reliability and measure excess loss above this floor. The shot-odds intervention has a different interpretation. Given the coordinates and rules, an exact solver can compute every conditional scoring probability, and rounding the supplied odds preserved every best shot. The extra table therefore has zero decision value for this solver. Gemini's increase from 12 to 57 solved pairs measures the benefit of having that calculation supplied.

\subsection{Does Gemini follow the cue when protection changes pieces?}
\label{sec:cue}
The 13 knight choices in Section~\ref{sec:fresh} could reflect a preference for the piece, uncertainty about tackle exposure, or an error deriving exposure from the board. To test whether Gemini can act on explicit evidence that favors either piece, we froze a third pass study after the 16-board comparison. On each of six fresh layouts, the prompt described two equally likely squares for a hidden pawn. The ordinary receiver was safe in one state and exposed in the other. A second receiver and a bishop were exposed in both states. The native engine verified the tackle geometry. A variant rule gave the designated protected receiver $0.75$ retention if tackled. Both other receivers lost possession if tackled. We exchanged the rook and knight between the ordinary and protected roles, keeping the geometry of the roles fixed.

The prompt supplied both possible tackle outcomes, a scout report about the ordinary receiver and its reliability. The scout was equally accurate in both states. After a \emph{safe} report of reliability $r$, the ordinary receiver's expected retention was $r$. After an \emph{exposed} report it was $1-r$. No realized hidden state was needed to score the decision. The bank crossed two reports, two reliabilities and two role assignments on every layout, covering all six menu orders across layouts. Gemini received one text-only request per cell, 48 total, in fresh contexts with the same generation settings as Section~\ref{sec:fresh}. Matched reliability prompts differed only in the printed percentage.

\begin{table}[htbp]
\centering
\caption{Gemini's choices on the six layouts in each cell of the cue experiment. The protected receiver retained possession with probability $0.75$ in all cells. Each listed choice maximized expected retention.}
\label{tab:cue}
\small
\begin{tabular}{llrll}
\toprule
Scout report & Reliability & Ordinary retention & Knight protected & Rook protected\\
\midrule
Safe & 90\% & $0.90$ & Rook 6/6 & Knight 6/6\\
Safe & 60\% & $0.60$ & Knight 6/6 & Rook 6/6\\
Exposed & 90\% & $0.10$ & Knight 6/6 & Rook 6/6\\
Exposed & 60\% & $0.40$ & Knight 6/6 & Rook 6/6\\
\bottomrule
\end{tabular}
\end{table}

The primary endpoint required both choices on the same layout and protection assignment. The model had to pass to the ordinary receiver on a 90\% safe report and to the protected receiver on a 60\% safe report. Gemini made all 12 reversals, six under each protection assignment, and chose protection on all 24 exposed-report controls (Table~\ref{tab:cue}). A policy always choosing protection would score 36/48 decisions and zero reversals. Gemini chose 24 knights and 24 rooks, maximizing expected retention on all 48 decisions. The 90\% cue raised attainable retention from $0.75$ without a report to $(0.90+0.75)/2=0.825$ before seeing which report arrived. The 60\% cue had zero decision value because both reports favored protection. Gemini attained the resulting mean of 78.75\% across the balanced prompts. Computing a legal tackle map from coordinates and estimating cue reliability from experience remain separate tests.

\section{Discussion}
On the original coordinate-only banks, Gemini changed its shot about as often on a stay as on a switch, yet changed its pass on nearly every switch and held its choice on every stay. With shot odds supplied, its overall change count stayed at 33 while the changes aligned much more closely with the required switches. The fresh passing boards exposed a different pattern. When only four of 16 boards favored the knight, all nine errors still went to that receiver. On the cue task it switched away from either piece whenever the stated probabilities warranted it. A fixed knight preference across these settings is inconsistent with the choices. Whether a missed tackle relation accounts for errors on the original full boards remains open.

The two geometry rules show that much of each task is recoverable from the public representation. They also suggest a hypothesis for the difference between tasks. Pass retention depends on a small set of discrete capture rules. Shot choice requires comparing graded payoffs across several possible goalkeeper destinations. The experiments changed the decision, objective, bank and arithmetic together. They do not identify which difference accounts for the gap or how Gemini arrived at its choices.

\paragraph{Locating the remaining difficulty.}
A rule choosing the ordinary receiver only for a 90\% safe report reproduces all 48 choices. The two reliability levels and one response per cell leave open whether the model computed a posterior or used a simpler comparison. The full-board and cue experiments differ in layout, protection rules, visibility and the information supplied about tackles. To isolate the cost of deriving exposure, a matched test could keep the original full board, receiver menu and tackle rules fixed. Add only a list of legal tacklers for each receiver in a second condition. Improvement would measure the benefit of supplying the relation between defender and receiver. Errors that persist with a correct tackle map would motivate testing how the model converts exposure into a choice.

\paragraph{What the controls leave unresolved.}
The shot bank deliberately selects near-goal decisions with a clear margin. Queens take 118 of the 120 shots, all 1.5 or 2.5 ranks from the target. Edit type and optimality are associated: 21 of 30 switches are keeper moves, and 21 of 30 stable edits are pawn moves. A reader who saw both boards and guessed ``switch on a keeper edit'' would classify 42/60 edits correctly without choosing a shot. The set of legal goalkeeper replies is unchanged on one switch pair and eleven stay pairs. Comparisons by mover or by menu change therefore cannot isolate independent causal effects. The geometry rule and behavioral diagnostic remain exploratory on this bank. The original passing bank uses one seed, a fixed hold after a successful pass, and only three distinct retention probabilities. The tackle-shape rule identifies the best receiver on 78 of the 80 boards. The two tasks test related decisions within one game and provide no estimate of performance in full matches. One response per input leaves repeated-call variability unmeasured. Under an illustrative independent-error calculation using Gemini's board-only accuracy, the expected solved-pair count is $60(52/120)^2=11.3$, close to the 12 observed. Aggregate pair success alone therefore provides limited evidence of additional structure in its errors. Repeated identical inputs would help distinguish response variability from stable sensitivity to the edited positions. The shot intervals describe variation among sampled pairs conditional on the three seeds and the recorded runs. Training exposure is unknown.

\paragraph{What a one-square edit can measure.}
In a chess middlegame, moving a blocker can open a line of attack, but the effect on winning usually depends on later play. In backgammon, the dice probabilities are known, yet valuing a move for its effect on winning still requires evaluating continuations~\cite{antonoglou}. Football Chess offers a shorter question with a tactical interpretation. In pair \texttt{m02}, a pass reaches the rook on either board. Moving one pawn exposes the rook to a tackle, dropping its retention from one to zero. We can measure the cost of overlooking that pawn for every offered pass, including those never played.

\paragraph{From the overhead view to the passer's view.}
Which defender is worth looking for before releasing the ball? Suppose the passer knows the pair, its two boards are equally likely, and the pawn's square is hidden from both the image and the roster. A perfect look raises expected retention by 25 percentage points on each of the four fresh switch pairs and by zero on the four stays (Section~\ref{sec:information}). At a positive cost below 25 points, an optimal passer would request the look only on the switch pairs. This gives a prediction for a future experiment in which looking is itself an action.

The companion Football Tactical LLM System~\cite{chen} estimates player positions under simulated visibility and separately classifies events at supplied actor and query times. Its tracking baseline leaves possession unknown, and its receiver study uses the next annotated actor as a proxy. A first exo--ego test could fix the passer and the instant before ball contact, show only tracks within the passer's estimated field of view, then reveal one additional defender from the overhead record. We could measure when that reveal changes the chosen pass. Tracking geometry establishes possible visibility. Attention would need independent evidence. Valuing an unchosen real pass would also require an estimate with stated uncertainty~\cite{soccermap}. Synchronized first- and third-person footage~\cite{egoexo} could support this test once the views are aligned. The companion project's video-to-pitch mapping has yet to pass global validation.

\paragraph{Code and data availability.}
The public \href{https://yuanrui3.github.io/Football-chess/benchmarks/paired-positions/}{benchmark page} links the primary datasets and citation metadata. The \href{https://github.com/yuanrui3/Football-chess/tree/8e87600e949d23675a293414391c54e63cbc64aa/benchmarks/paired-positions}{paired-position benchmark} is available under \path{benchmarks/paired-positions/} at commit \texttt{8e87600} of \url{https://github.com/yuanrui3/Football-chess}. The \benchmarkfile{positions.json}{shot data} and \benchmarkfile{pass/data.json}{pass data} contain the boards, action menus, engine values and decoded model choices. The exact prompts appear in a \benchmarkfile{prompts.json}{separate shot prompt file} and in the pass data.

From the benchmark directory or the extracted paper source, run \benchmarkfile{score.mjs}{\texttt{node anc/score.mjs}} to reproduce the shot results and \benchmarkfile{geometry.mjs}{geometry baseline}. Run \benchmarkfile{pass/reproduce.mjs}{\texttt{node anc/pass/reproduce.mjs}} to reproduce the pass results and independently check the retention values. Run \benchmarkfile{native/verify.mjs}{\texttt{node anc/native/verify.mjs}} to replay both banks with the \benchmarkfile{native/engine.mjs}{frozen rules-version-22 engine}. All three commands work offline with Node.js 24.2 or later.

The earlier visual walkthrough and scorer are in \path{anc/walkthrough/}. For Section~\ref{sec:fresh}, \path{anc/followup/} contains the new boards, exact quiz images and prompts, decoded choices, and freeze record. From the extracted source, \texttt{node anc/followup/reproduce.mjs} regenerates the bank and scores the choices. The post hoc calculation in Section~\ref{sec:information} can be repeated with \texttt{node anc/followup/value-of-information.mjs}. For the cue experiment, \texttt{node anc/cue/reproduce.mjs} checks the frozen inputs and reproduces the matched decisions in Table~\ref{tab:cue}.

The exo--ego code is at \url{https://github.com/yuanrui3/Ego-exo-football}. Reference~\cite{chen} links to the README that describes the tracking system.

\paragraph{Tool use.}
OpenAI Codex assisted with experiment implementation, analysis, manuscript drafting and reference checking. Gemini 4 Neon provided editorial feedback and, at its High thinking setting, reviewed the manuscript text and the revised cue results for interpretability. Claude Fable 5.2 at Extra effort reviewed a summary of the design, results and reproducibility for submission readiness and discussed competing explanations and follow-up tests for the observed receiver choices. The evaluated models answered in separate inference calls under the protocols described above.

\appendix
\section{Game rules relevant to the benchmark}
\label{app:rules}
This appendix describes the rules used to evaluate the offered shots and passes in engine rules version 22. The constants define a board game. They were chosen as game parameters and have no fitted interpretation as probabilities for real football. The ancillary prompts contain the instructions shown to the evaluated models.

\subsection{Pitch and possession}
Files a--n run left to right and ranks 1--17 run bottom to top. White attacks upward and defends the bottom end; Black attacks downward and defends the top. The goal occupies files f--i beyond either end. The penalty boxes occupy files e--j at ranks 1--3 and 15--17. Each occupied square contains a player identified by team and piece type: keeper (K), queen (Q), rook (R), bishop (B), knight (N) or pawn (P). The ball carrier has possession, and the other team defends. Players obstruct paths according to the rules below. Any square absent from the supplied roster is empty.

Outfield players may occupy the boxes, except the goalmouth squares f--i on ranks 1 and 17. A keeper may enter its own box and may leave it, but cannot enter the opposing box. Holding leaves the player on its current square. Each board is an independent decision state. The changed square in a pair constructs an alternative state and does not prescribe a legal move to be played between the two evaluations.

\subsection{Shots and the goalkeeper reply}
Every shot position starts with the defending keeper inside its own box and offers two target files. The attacker declares one target or chooses disguise, which draws the two targets with equal probability. After seeing a declaration, the keeper chooses a reply knowing the target. Against disguise it commits to the reply before the draw. Replies comprise holding and one-square horizontal, vertical or diagonal steps to empty on-board squares that respect the keeper's box restriction. It may step out of its box. The reply depends on occupancy and these rules. Chess check and attacked-square maps impose no additional restriction.

A queen's shot follows the segment between the shooter center and the target center on the goal line. Each opposing body whose square interior the segment crosses multiplies the goal probability by 0.55. Touching only a corner or running along an edge does not cross the interior. A knight's shot ignores these blockers. A keeper inside its box uses save coverage. Outside its box it contributes to the blocking bodies. The target is half a rank beyond the board, at rank 17.5 for White and 0.5 for Black.

The task ends at the immediate goal outcome. The optimal shot maximizes the scoring probability against the keeper's most effective reply, as defined in Equation~\ref{eq:value}.

\subsection{Numerical scoring probability}
\label{app:scoring}
These equations transcribe the reference formula supplied as plain text in every shot prompt for the queen and knight carriers in this bank. Independently computing this formula recovered all 360 native action values with maximum absolute probability error below $5.1\times10^{-16}$. Coordinates use $x=0,\ldots,13$ for files a--n and the printed rank for $y$. For shooter $S$, target $T$ and the keeper's chosen position $K$, define
\[
 C(z)=\min(1,\max(0,z)),\quad L=\|T-S\|_2,\quad
 u=(T-S)/L,\quad d=|T_y-S_y|.
\]
The opening between the physical posts, in radians, and its reference width are
\[
 w=\arctan\frac{8.5-S_x}{d}-\arctan\frac{4.5-S_x}{d},
 \qquad w_0=2\arctan(2/d),\qquad
 h=C\left(\frac{180w/\pi-45}{45}\right).
\]
With the keeper inside its box, the on-target factor $A$ is obtained by clamping the following product to $[0.01,0.98]$:
\begin{align*}
 A_{\mathrm{raw}}={}&0.98(1.15)
 \exp[-0.018L-0.06\max(0,L-3)^2]\\
 &{}\times(0.65+0.35d/L)
 [0.1+0.9C(w/w_0)^{1.5}]\\
 &{}\times[1+0.25h^2(3-2h)]\,c_S .
\end{align*}
Here $c_S=1.15$ if the shooter is on file g or h and $c_S=1$ otherwise. To calculate the save factor, set
\begin{align*}
 v&=(K-S)\cdot u,\qquad \ell=\min(L,\max(0,v)),\\
 c&=\|K-(S+\ell u)\|_2,\qquad \tau=\ell/6,\qquad
 b=|K_y-y_{\mathrm{end}}|,\\
 \rho&=0.4+1.3\max(0,\tau-0.15),\\
 \eta&=0.9\{1-\exp[-(0.35+\tau+b|u_y|/6)/0.45]\},
\end{align*}
where $y_{\mathrm{end}}=17$ for a White attack and $1$ for a Black attack. The dive and rush components are
\begin{align*}
 D&=\begin{cases}\eta\exp[-(c/\rho)^2],&v\geq0,\\0,&v<0,\end{cases}\\
 H&=\begin{cases}
 0.2C(b/2)\max(0,1-\|K-S\|_2^2/36)\exp[-(c/0.9)^2],
       &0<v<L,\\
 0,&\text{otherwise},
 \end{cases}\\
 \mathrm{Save}&=\min(0.9,\max(0,D+(1-D)H)).
\end{align*}
When the keeper replies outside its box, $\mathrm{Save}=0$ and $A$ becomes the ordinary pass-accuracy factor
\[
 A=\max(0.25,1-0.1\max(0,n-10)),\qquad
 n=\max(|T_x-S_x|,|y_{\mathrm{kick}}-S_y|),
\]
where $y_{\mathrm{kick}}=18$ for White and $0$ for Black. This distance uses the integer kick destination, while the blocker segment ends at the half-rank goal line. Finally,
\[
 q(K,T)=A(1-\mathrm{Save})\,0.55^m .
\]
The count $m$ includes opposing bodies crossing the segment, with the keeper placed at its reply square and included only when outside its box. For a knight carrier $m=0$. Disguise averages the two target probabilities at the same $K$ before minimizing across replies.

\subsection{Passing and the immediate defensive challenge}
\label{app:passing}
Every passing question offers three ordinary queen passes to named teammates on unobstructed queen rays, two to six squares away. Each pass is legal and onside, and its accuracy is one. The ball transfers to the chosen receiver without moving the players. In normal play, positional actions follow the kick. For this study the attack ends that phase by holding. The defender then has one ordinary move or hold, and the evaluation ends. A move that wins possession of the ball determines the retention probability. Fouls are disabled.

The following shapes determine whether a defender can tackle the receiver from its starting square. They apply equally in both attacking directions.
\begin{center}
\begin{tabular}{lp{3.95in}}
\toprule
Defender & Tackle reach\\
\midrule
Queen & Same file, rank or diagonal, up to five squares.\\
Rook & Same file or rank, up to five squares.\\
Bishop & Same diagonal, up to five squares.\\
Knight & A displacement of $(1,2)$ or $(2,1)$ in absolute file and rank differences.\\
Pawn & One diagonal step in any direction.\\
Keeper & One neighboring square, including diagonals.\\
\bottomrule
\end{tabular}
\end{center}
Sliding tackle rays require clear intervening squares and respect the defender's square restrictions. Knights jump intervening pieces. A tackle additionally requires a legal landing square adjacent to the receiver, at most ten squares from the defender's start when distance is the larger of the file and rank differences. This square must be empty or the defender's own starting square and must respect its occupancy restrictions. The receiver's square is occupied, so the tackler lands beside it.

Any legal tackle wins possession with probability one against a non-knight receiver and one-half against a knight. Failure leaves possession with the receiver. Thus an untacklable receiver retains possession with probability one, an exposed knight with one-half, and any other exposed receiver with zero. The defensive reply maximizes the probability of winning possession. The answer is the receiver maximizing retention after that reply. The supplied pass legality is part of the question: the model compares the three offered recipients using their resulting exposure. The native verification checked all 1,640 legal tackle landings over the 240 offered passes.

\paragraph{Worked pass calculation.}
In Figure~\ref{fig:pass}, passing to the bishop on i15 in A leaves it three files from the black rook on l15 along a clear rank. The rook has a legal tackle, so retention is $1-1=0$. The white rook on g9 can be tackled by the black queen on e9 and also retains with probability zero. The white knight on i7 can be tackled by the black bishop on f4, but survives with probability $1-0.5=0.5$. Passing to the knight is therefore optimal. In B, the defending rook is on l14, which shares neither file nor rank with i15. No other defender can tackle that receiver, so its retention rises to one. The other receiver values remain zero and one-half. Passing to the bishop becomes optimal.
\clearpage
\section{A visual walkthrough of four pass pairs}
\label{app:walkthrough}

After reviewing the study results, the sole author completed an eight-position quiz drawn from the pass bank. A fixed seed selected one pair from each combination of attacking color and switch or stay, using pair identifiers independently of the recorded choices. Selection excluded the worked example in Figure~\ref{fig:pass}. The quiz showed one pitch diagram at a time, with the three receivers numbered in the original menu order. Three other questions separated the two members of each pair. The author could consult the rules and revisit earlier choices.

Before submitting, the author asked whether the receiver could move and whether to select the knight when every receiver could be tackled. Codex confirmed that the attack holds, an exposed knight retains possession half the time, and a receiver beyond any legal tackle retains possession with certainty. Both clarifications applied to the quiz as a whole.

\begin{table}[htbp]
\centering
\caption{Pass choices on the four quiz pairs. Both the author and Gemini selected the receivers shown. Parentheses give the probability of retaining possession after the defender's best reply.}
\label{tab:walkthrough}
\small
\begin{tabular}{llll}
\toprule
Questions & Defender edit & First receiver & Second receiver\\
\midrule
1 and 5 & Bishop b10 to a10 & Knight f10 (0.5) & Bishop d8 (1)\\
2 and 6 & Bishop b11 to b10 & Knight f14 (1)   & Knight f14 (0.5)\\
3 and 7 & Pawn l13 to l14   & Knight e10 (0.5) & Rook k14 (1)\\
4 and 8 & Bishop h15 to h14 & Knight f12 (1)   & Knight f12 (0.5)\\
\bottomrule
\end{tabular}
\end{table}

The author chose the optimal receiver on all eight boards and solved all four pairs. On the two switch pairs the author left the knight for a newly safe bishop or rook (Table~\ref{tab:walkthrough}). Always selecting the knight would have scored six boards and two pairs. Rescoring Gemini's original coordinate-only responses gave the same eight choices as the author.

This walkthrough documents how one reader familiar with the study used a diagram and clarified rules. The author and Gemini saw different displays and had different opportunities for clarification. Estimating human performance or comparing it with model accuracy would require independent participants and matched conditions.

\begin{thebibliography}{99}
\bibitem{gardner} M. Gardner et al. \href{https://aclanthology.org/2020.findings-emnlp.117/}{Evaluating Models' Local Decision Boundaries via Contrast Sets}. Findings of EMNLP, 1307--1323, 2020.
\bibitem{ribeiro} M. T. Ribeiro, T. Wu, C. Guestrin, and S. Singh. \href{https://aclanthology.org/2020.acl-main.442/}{Beyond Accuracy: Behavioral Testing of NLP Models with CheckList}. ACL, 4902--4912, 2020.
\bibitem{costarelli} A. Costarelli et al. \href{https://arxiv.org/abs/2406.06613}{GameBench: Evaluating Strategic Reasoning Abilities of LLM Agents}. arXiv:2406.06613, 2024.
\bibitem{wang} S. Wang et al. \href{https://aclanthology.org/2025.naacl-short.52/}{Explore the Reasoning Capability of LLMs in the Chess Testbed}. NAACL, 611--622, 2025.
\bibitem{jain} O. Jain and D. Kumar. \href{https://arxiv.org/abs/2604.05681}{LudoBench: Evaluating LLM Behavioural Decision-Making Through Spot-Based Board Game Scenarios in Ludo}. arXiv:2604.05681, 2026.
\bibitem{holm} S. Holm. \href{https://www.jstor.org/stable/4615733}{A Simple Sequentially Rejective Multiple Test Procedure}. Scandinavian Journal of Statistics, 6(2), 65--70, 1979.
\bibitem{antonoglou} I. Antonoglou, J. Schrittwieser, S. Ozair, T. Hubert, and D. Silver. \href{https://openreview.net/pdf?id=X6D9bAHhBQ1}{Planning in Stochastic Environments with a Learned Model}. ICLR, 2022.
\bibitem{howard} R. A. Howard. \href{https://doi.org/10.1109/TSSC.1966.300074}{Information Value Theory}. IEEE Transactions on Systems Science and Cybernetics, 2(1), 22--26, 1966.
\bibitem{egoexo} K. Grauman et al. \href{https://ego-exo4d-data.org/}{Ego-Exo4D: Understanding Skilled Human Activity from First- and Third-Person Perspectives}. CVPR, 2024.
\bibitem{chen} Y. Chen. \href{https://github.com/yuanrui3/Ego-exo-football/blob/f427cbab73d76830b081ce8d782f329458d546d2/README.md}{Football Tactical LLM System}. Research software, researcher-handover README, snapshot \texttt{f427cba}, November 2026.
\bibitem{soccermap} J. Fern\'andez and L. Bornn. \href{https://arxiv.org/abs/2010.10202}{SoccerMap: A Deep Learning Architecture for Visually-Interpretable Analysis in Soccer}. ECML PKDD, 491--506, 2020.
\end{thebibliography}
\end{document}